\documentclass[10 pt,twocolumn,notitlepage]{article}
\usepackage[american]{babel}
\usepackage[strings]{underscore}
 \usepackage{ amsmath} 
\usepackage{amssymb }
\usepackage{bbm}
\usepackage{graphicx}
\usepackage{url}
\usepackage[left=2cm,right=2cm,top=2cm,bottom=2cm]{geometry}

\usepackage{natbib} 

\title{On the Convergence of Locally Adaptive and Scalable Diffusion-Based Sampling Methods for Deep Bayesian Neural Network Posteriors}

\author{Tim Rensmeyer, Oliver Niggemann}

\date{}
  
\begin{document}
\maketitle
\begin{abstract}
Achieving robust uncertainty quantification for deep neural networks represents an important requirement in many real-world applications of deep learning such as medical imaging where it is necessary to assess the reliability of a neural network's prediction.
Bayesian neural networks are a promising approach for modeling uncertainties in deep neural networks. Unfortunately, generating samples from the posterior distribution of neural networks is a major challenge. One significant advance in that direction would be the incorporation of adaptive step sizes, similar to modern neural network optimizers, into Monte Carlo Markov chain sampling algorithms without significantly increasing computational demand.
Over the past years, several papers have introduced sampling algorithms with claims that they achieve this property. However, do they indeed converge to the correct distribution?
In this paper, we demonstrate that these methods can have a substantial bias in the distribution they sample, even in the limit of vanishing step sizes and at full batch size. 
\end{abstract}

\section{Introduction}
Since the beginning of the deep learning revolution, neural networks have achieved impressive results in many domains such as vision \citep{ImageNet,ResNet,YOLO,UNET}, language \citep{Transformer, BERT, GPT} and Markovian decision processes \citep{DQL,AlphaZero,DPG}. However, despite their successes the adoption of deep learning models in safety-critical domains has been slow, in large part due to a lack of interpretability and limited robustness to outliers \citep{Safety}. Another desirable property in safety-critical applications is a reliable uncertainty quantification for neural network predictions to assess which neural network predictions are reliable and which are likely to be wrong \citep{Uncertainty}.\\
Bayesian Neural Networks (BNNs) are a promising approach to improve robustness to outliers while also achieving high-quality uncertainty quantification \citep{BNNRobust,BNNquality,BNNSurvey}. Unfortunately, sampling from the Bayesian posterior over neural network parameters represents a challenging task. The gold-standard approach to sampling a distribution close to the true posterior is via Monte Carlo Markov Chain (MCMC) methods, where a stochastic process over the neural network parameters is simulated, which is known to converge in distribution to the true posterior for appropriate learning rate schedules. For these approaches, a fairly complete framework of possible Markov chains that converge to the correct distribution has been established by \citet{CompleteRecipe}. However, a key component of modern (classical) neural network optimizers is the usage of adaptive step sizes \citep{Optimizers} and the results established by \citet{CompleteRecipe} appeared to imply that using adaptive step sizes analogously in MCMC methods requires the computationally costly calculation of an additional term. The evaluation of this term will typically require the calculation of the second-order derivatives of the neural network predictions w. r. t.  the neural network parameters. This approximately doubles the size of the computational graph and consequently also approximately doubles the computational time and memory requirements of each optimization step. To circumvent this problem, several alternative algorithms have been proposed, where the calculation of such a pathological term is either entirely unnecessary or where its neglection only causes a minor deviation to the distribution that the process converges to. \\
However, these results appear to directly contradict some principles from the ergodic theory of Stochastic Differential Equations (SDEs). 
Consequently, the main research question of this paper is the following:\\
Do the proposed algorithms indeed converge to a distribution close to the true posterior?\\
Furthermore, we investigate the following more general research question:
What is the consequence of using adaptive step sizes without including a correction term to the dynamics?

\section{Background: Bayesian Neural Networks} 
Throughout this paper, we make the conventions that bold case symbols represent vector-valued quantities and $\boldsymbol{\theta}$ is a vector containing all the trainable parameters of a neural network e.g. weights and biases.\\

The main difference between Bayesian neural networks and classical ones is that for BNNs the trainable parameters are modeled probabilistically. The starting point for BNNs is a prior probability density $p(\boldsymbol{\theta})$, that expresses preexisting knowledge of which neural network parameters are more likely to result in a good model of the data distribution the neural network is supposed to model. This preexisting knowledge is then refined via a training dataset $\mathcal{D}=\{s_1, ..., s_M\}$ by using Bayes rule:
$$p(\boldsymbol{\theta}|\mathcal{D})=\frac{p(\mathcal{D}|\boldsymbol{\theta})p(\boldsymbol{\theta})}{p(\mathcal{D})}.$$
A prediction on a new data point $s^*$ can then be made via
$$p(s^*|\mathcal{D})=\int p(s^*|\boldsymbol{\theta})p(\boldsymbol{\theta}|\mathcal{D})d\boldsymbol{\theta}=\mathbb{E}_{p(\boldsymbol{\theta}|\mathcal{D})}[ p(s^*|\boldsymbol{\theta})].$$
While this expectation w. r. t. to posterior density is almost always intractable for neural network models, it can be evaluated through the law of large numbers as 
$$\mathbb{E}_{p(\boldsymbol{\theta}|\mathcal{D})}[ p(s^*|\boldsymbol{\theta})]\approx \frac{1}{N}\sum_{k=1}^N p(s^*|\boldsymbol{\theta}_k)$$
where $\boldsymbol{\theta}_k , k\in \{1, ..., N\}$ are independent samples from the posterior distribution. 
These samples can be generated by simulating a stochastic process over the neural network parameters, which is known to converge in distribution to the posterior. A short overview of relevant processes is given in the following subsections. A more in-depth discussion of BNNs can be found in the work by \citet{BNNSurvey}. 

\subsection{Stochastic Gradient Langevin Dynamics}
One of the most popular algorithms for sampling the posterior distribution of neural networks is Stochastic Gradient Langevin Dynamics (SDLD) \citep{SGLD}.
This algorithm is based on a time-discretized simulation of the Stochastic Differential Equation (SDE)

$$d\boldsymbol{\theta}_t =\frac{1}{2} \nabla_{\boldsymbol{\theta}_t}u(\boldsymbol{\theta}_t)dt +d\boldsymbol{B}_t,$$
where $u(\boldsymbol{\theta}_t)=\log(p(\boldsymbol{\theta}_t|\mathcal{D}))$ and $d\boldsymbol{B}_t$ are the increments of a Brownian motion.
These diffusions provably converge to the distribution $p(\boldsymbol{\theta}|\mathcal{D})$. Unfortunately, objective functions such as $u(\boldsymbol{\theta}_t)$ will oftentimes have pathological curvature for deeper neural network architectures characterized by vanishing gradients for many of the neural network parameters. This can slow down the convergence of both classical gradient descent-based optimization and MCMC sampling methods for deep neural networks to the point where no convergence can be achieved in a practical amount of time. 
Modern neural network optimizers get around this problem by employing an individual adaptive step size for each parameter which is roughly inversely proportional to the current magnitude of its derivative \citep{Optimizers}.
 A major breakthrough for deep neural network-focused MCMC methods was the introduction of Stochastic Gradient Riemannian Langevin Dynamics (SGRLD) \citep{SGRLD} which can use adaptive step sizes similar to modern neural network optimizers by introducing a Riemannian metric $G(\boldsymbol{\theta})$ over the space of neural network parameters.
The resulting stochastic differential equation is
$$ d\boldsymbol{\theta}_t = \frac{1}{2}G(\boldsymbol{\theta}_t)\nabla_{\boldsymbol{\theta}_t}u(\boldsymbol{\theta}_t)dt +\frac{1}{2}\boldsymbol{\Gamma}(\boldsymbol{\theta}_t)dt + G(\boldsymbol{\theta}_t)^{\frac{1}{2}}d\boldsymbol{B}_t,$$
which still converges to the posterior but allows for adaptive step sizes via the metric $G(\boldsymbol{\theta})$. One example metric is the diagonal metric
$$G(\boldsymbol{\theta})=diag\left(\frac{1}{\lambda +\sqrt{ \nabla_{\boldsymbol{\theta}}u(\boldsymbol{\theta})\odot \nabla_{\boldsymbol{\theta}}u(\boldsymbol{\theta})}}\right)$$ where $\lambda$ is a stability constant, $\odot$ is the element-wise product and the square root and division are applied element-wise. The drawback of this new equation lies in the introduction of a new term $$\boldsymbol{\Gamma}_i(\boldsymbol{\theta}) := \sum_j \frac{\partial G_{i,j}(\boldsymbol{\theta})}{\partial \boldsymbol{\theta}_j}.$$ Because the metric will typically contain the gradient of the log posterior since it is supposed to increase step sizes when those gradients get small, the evaluation of $\boldsymbol{\Gamma}$ usually requires the computation of second-order derivatives of the log posterior with respect to the parameters. This effectively doubles the size of the computational graph and hence also the required computation time and GPU memory of each time step. Later a very general framework for incorporating adaptive step sizes in sampling methods was discovered \citep{CompleteRecipe}. Unfortunately, it also required the computationally intensive calculation of additional terms each time step.
As a consequence, some authors have tried to develop alternative algorithms, where such terms are not required or the neglection causes only a small change in the distribution the algorithm converges to.

\subsection{Proposed Alternatives}
In this subsection, we will introduce the algorithms, which incorporate adaptive step sizes without a costly correction term and for which the authors claim, that the process will converge to almost the correct distribution. We discuss here only the full-batch variants of the proposed algorithms. The mini-batch versions can be found in the original works.
By far the most popular one is the Preconditioned Stochastic Gradient Langevin Dynamics (PSGLD) algorithm \citep{PSGLD} which at the time of writing has over 350 citations. The algorithm is given by the updates
$$\boldsymbol{V}_{t} = \alpha \boldsymbol{V}_{t-\epsilon}+(1-\alpha) \nabla_{\boldsymbol{\theta}_t}u(\boldsymbol{\theta}_t)\odot \nabla_{\boldsymbol{\theta}_t}u(\boldsymbol{\theta_t})$$
$$G_t=diag\left(\frac{1}{\lambda + \sqrt{\boldsymbol{V}_t}}\right)$$
$$\boldsymbol{\theta}_{t+\epsilon}=\boldsymbol{\theta}_{t}+\frac{\epsilon}{2}\left(G_t\nabla_{\boldsymbol{\theta}_t}u(\boldsymbol{\theta}_t) + \boldsymbol{\Gamma}(\boldsymbol{\theta}_t)\right) +G_t^{\frac{1}{2}}\boldsymbol{\mathcal{N}}_t(0,\epsilon \mathbb{I}).$$
Here, $\epsilon$ is the step size, $\alpha$ is a hyperparameter typically slightly smaller than one, $\lambda$ is a stability constant close to zero, $ \mathbb{I}$ is the identity matrix and $\boldsymbol{\mathcal{N}}_t(0,\epsilon \mathbb{I}), t\in \mathbb{N}_0$ are independent Gaussian random variables $\sim \boldsymbol{N}(0,\epsilon \mathbb{I})$. Again the square root and division are applied element-wise.\\
Two other similar algorithms were recently introduced called SGRLD in Monge metric and SGRLD in Shampoo metric \citep{scalable}. SGRLD in Monge metric is similar to PSGLD but changes the definition of $\boldsymbol{V}_t$ and $G_t$ to
$$\boldsymbol{V}_{t} = \alpha \boldsymbol{V}_{t-\epsilon}+(1-\alpha) \nabla_{\boldsymbol{\theta}_t}u(\boldsymbol{\theta}_t),$$
$$G_t= \mathbb{I} -\frac{\beta ^2}{1+\beta ^2||\boldsymbol{V}_{t}||^2}\boldsymbol{V}_{t}\boldsymbol{V}_{t}^T,$$
where $\beta ^2$ is a hyperparameter.
The construction of the Shampoo metric is quite involved but luckily in the case of one-dimensional distributions, the resulting algorithm coincides with PSGLD where the stability constant $\lambda$ is set to zero. Since this is the only case we will analyze in the next section we do not introduce the general algorithm here. The interested reader can find the general definition in the work by \citet{scalable}. 

  In all three algorithms, the authors claim that dropping $\boldsymbol{\Gamma}(\boldsymbol{\theta}_t)$ will not change the distribution that the process converges to significantly.\\
Finally, there is the Adam SGLD algorithm \citep{AdamSGLD} which is defined via 
$$\boldsymbol{m}_t=\beta \boldsymbol{m}_{t-\epsilon} +(1-\beta)\nabla_{\boldsymbol{\theta}_t}u(\boldsymbol{\theta}_t),$$
$$\boldsymbol{\theta}_{t+\epsilon}=\boldsymbol{\theta}_{t}+\frac{\epsilon}{2}(\nabla_{\boldsymbol{\theta}_t}u(\boldsymbol{\theta}_t) +aG_t\boldsymbol{m}_t)+\boldsymbol{\mathcal{N}}_t(0,\epsilon \mathbb{I})$$
where $G_t$ coincides with the definition of the PSGLD algorithm, $\beta$ is another hyperparameter that is typically slightly smaller than one, and $a>0$ is a hyperparameter controlling the additional drift term. Again the authors claim, that this is a suitable algorithm for sampling $p(\boldsymbol{\theta}|\mathcal{D})$.

\section{Convergence Analysis of Sampling Algorithms with Adaptive Step Sizes}
The main working principle of all SDE-based sampling approaches is that they converge to a stationary density $\pi(\boldsymbol{\theta})$ which should be equal to the target density one tries to sample, e.g. the posterior over the neural network parameters in this case. This density is invariant under the SDE \citep{CompleteRecipe}, e.g. if at $t=0$ the variable $\boldsymbol{\theta}_t$ is distributed according to $\pi$, then it will have this distribution for all $t>0$. In fact, it is not difficult to see that any distribution a process can converge to has to be a stationary distribution of the process. For a given SDE, $\pi$ can in principle be determined via solving the associated Fokker-Planck partial differential equation \citep{Oskendal}.
A fundamental property of the SDEs under consideration here is an ergodic property, that enables the computation of expectations of a function $f$ w. r. t. the stationary density as a time average of the function \citep{Borodin}
$$\mathbb{E}_{\pi}\left[f(\boldsymbol{\theta})\right]=\lim_{T\rightarrow \infty}\frac{1}{T}\int_{t=0}^T f(\boldsymbol{\theta}_t)dt.$$
When convergence speed and the effects of time discretization were investigated in the papers for the proposed algorithms, it was done using this property by analyzing the discrepancy of $\mathbb{E}_{\pi}\left[f(\boldsymbol{\theta})\right]$ and the time discretized analog of $\frac{1}{T}\int_{t=0}^T f(\boldsymbol{\theta}_t)dt$ in the large $T$ limit. 

Unfortunately, this ergodic property also seems to indicate that incorporating adaptive step sizes into the sampling procedure without a correction term should at best be very difficult and at worst simply not possible. This can be seen by applying the ergodic property to the indicator function $$\mathbbm{1}_U(\boldsymbol{\theta})=\left\{\begin{array}{lr}
        1, & \text{for } \boldsymbol{\theta} \in U\\
        0, & \text{otherwise}
        \end{array}\right\} 
.$$
This yields $\mathbb{P}_{\pi}(U)=\mathbb{E}_{\pi}\left[\mathbbm{1}_U(\boldsymbol{\theta})\right]=\lim_{t\rightarrow \infty}\frac{1}{T}\int_{t=0}^T \mathbbm{1}_U(\boldsymbol{\theta}_t)dt$
- when sampling from the stationary density the probability of sampling a parameter vector $\boldsymbol{\theta}$ from some region $U$ of parameter space is simply given by the fraction of time the process $\boldsymbol{\theta}_t$ spends in that region. If the step size is selectively doubled in some region $U$ of parameter space the process will roughly spend only half as much time in that region as before and hence the stationary density changes.

\subsection{The neccessity of the $\boldsymbol{\Gamma}$ Term in Riemannian Langevin Dynamics}
The majority of the proposed sampling methods we discuss here are based on Riemannian Langevin dynamics. In all of those examples, it is claimed that the pathological $\boldsymbol{\Gamma}$ term can be neglected since the overall term is supposedly very small. But based on the previous analysis this seems strange. In fact, it will be shown here that it is unlikely that the  $\boldsymbol{\Gamma}$ term can ever be neglected for any sensible adaptive metric $G(\boldsymbol{\theta})$ without significantly changing the stationary density. To see this one simply has to look at the one-dimensional case. In the case of a one dimensional SDE $d\theta_t=\mu(\theta_t)dt +\sigma(\theta_t)dB_t$ the stationary density $\pi(\theta)$ is known to be 
$$\pi(\theta)=\frac{Z}{\sigma(\theta)^2}\exp\left(2\int^{\theta}\frac{\mu(x)}{\sigma(x)^2}dx\right),$$
where $Z$ is a normalization constant \citep{Borodin}.\\
Applying this formula to one-dimensional Riemannian Langevin Dynamics under neglection of $\Gamma$:
$$ d\theta_t = \frac{1}{2}G(\theta_t)\frac{d}{d\theta_t}u(\theta_t)dt + G(\theta_t)^{\frac{1}{2}}d{B}_t,$$
results in  (see Appendix A.1 for a derivation)
$$\pi(\theta)=\frac{Z}{G(\theta)}p(\theta|\mathcal{D}).$$
Unless the metric $G$ has almost the same value for all $\theta$ where $p(\theta|\mathcal{D})$ is reasonably large, the stationary density $\pi$ will significantly deviate from $p(\theta|\mathcal{D})$. However, the main function of the metric is to increase step sizes when the parameter derivatives get small. Since the derivatives vanish at local maxima of the log-likelihood it therefore seems unlikely that this approach can ever work without significantly changing the stationary distribution.\\
So what goes wrong in the mentioned algorithms?\\
All the sampling methods based on Riemannian Langevin dynamics where the $\boldsymbol{\Gamma}$ term was claimed to be negligible
are not actually time discretizations of Riemannian Langevin dynamics. The main problem is, that in the limit of vanishing step sizes, the $\boldsymbol{\Gamma}$ term is much smaller than what would be dictated by Riemannian Langevin dynamics as will be shown in the following. \\
All the algorithms involve a valid metric $G(\boldsymbol{\theta}_t)=F(\boldsymbol{h}(\boldsymbol{\theta}_t))$ for which $\boldsymbol{\Gamma}(\boldsymbol{\theta}_t)$ is not negligible. However, instead of using the term $\boldsymbol{h}(\boldsymbol{\theta}_t)$ itself, which carries the entire dependence on $\boldsymbol{\theta}_t$ it is replaced by its exponentially moving average $\boldsymbol{V}_t=\alpha \boldsymbol{V}_{t-\epsilon} +(1-\alpha) \boldsymbol{h}(\boldsymbol{\theta}_t)$. For example in PSGLD the function $\boldsymbol{h}(\boldsymbol{\theta})$ is given by 
$\nabla_{\boldsymbol{\theta}}u(\boldsymbol{\theta}) \odot \nabla_{\boldsymbol{\theta}}u(\boldsymbol{\theta}) $.
From now on we will denote with $G^{\alpha,t}$ and $\boldsymbol{\Gamma}^{\alpha,t}$ the corresponding quantities where $\boldsymbol{h}(\boldsymbol{\theta}_t)$ has been replaced by its exponential moving average.\\
Now the only explicit dependence of the metric $G^{\alpha,t}$ on the current parameter vector $\boldsymbol{\theta}_t$ comes from the $(1-\alpha) \boldsymbol{h}(\boldsymbol{\theta}_t)$ term in the exponential moving average, where $(1-\alpha)$ will be close to zero for typical values of $\alpha$.
 Consequently, by introducing this exponential moving average almost all explicit dependence of $G^{\alpha,t}$ on the current parameter vector $\boldsymbol{\theta}_t$ is lost, even though implicitly $G^{\alpha,t}$ still depends strongly $\boldsymbol{\theta}_t$ due to the high correlation of $\boldsymbol{h}(\boldsymbol{\theta}_t)$ with previous values $\boldsymbol{h}(\boldsymbol{\theta}_{t-\epsilon}), \boldsymbol{h}(\boldsymbol{\theta}_{t-2\epsilon}), ...\;$.\\
However, because $\boldsymbol{\Gamma}^{\alpha,t}$ is constructed from the derivatives of $G^{\alpha,t}$ w.r.t. the current parameters $\boldsymbol{\theta}_t$ and the explicit dependence of $G^{\alpha,t}$ on the current parameter vector is very weak, they find that $\boldsymbol{\Gamma}^{\alpha,t}=\mathcal{O}(1-\alpha)$ and is hence negligible for $\alpha$ sufficiently close to $1$. 
But proceeding like this is simply not something that you can do in Riemannian Langevin dynamics where the metric is supposed to depend only on the current parameter vector.
The main problem is, that in the limit of vanishing step sizes $\epsilon$ we have 
$\boldsymbol{V}_t \rightarrow \boldsymbol{h}(\boldsymbol{\theta}_t)$ and hence $G^{\alpha, t} \rightarrow G(\boldsymbol{\theta}_t)$.
But $\frac{\partial }{\partial \boldsymbol{\theta}_{t,j}} \boldsymbol{V}_{t,l} =(1-\alpha) \frac{\partial }{\partial \boldsymbol{\theta}_{t,j}} \boldsymbol{h}_l(\boldsymbol{\theta}_t)$
and consequently $$\boldsymbol{\Gamma}^{\alpha, t}_i := \sum_j \frac{\partial G^{\alpha,t}_{i,j}}{\partial \boldsymbol{\theta}_{t,j}}=\sum_{j,l} \frac{\partial \boldsymbol{V}_{t,l}}{\partial \boldsymbol{\theta}_{t,j}}\frac{\partial G^{\alpha,t}_{i,j}}{\partial \boldsymbol{V}_{t,l}}$$
$$=\sum_{j,l} (1-\alpha)\frac{\partial \boldsymbol{h}_l(\boldsymbol{\theta}_t)}{\partial \boldsymbol{\theta}_{t,j}}\frac{\partial G^{\alpha,t}_{i,j}}{\partial \boldsymbol{V}_{t,l}}$$
$$ \rightarrow \sum_{j,l} (1-\alpha)\frac{\partial \boldsymbol{h}_l(\boldsymbol{\theta}_t)}{\partial \boldsymbol{\theta}_{t,j}}\frac{\partial G_{i,j}}{\partial \boldsymbol{h}_l(\boldsymbol{\theta}_t)}$$
$$=\sum_{j} (1-\alpha)\frac{\partial G_{i,j}(\boldsymbol{\theta}_t)}{\partial \boldsymbol{\theta}_{t,j}}=(1-\alpha)\boldsymbol{\Gamma}_i(\boldsymbol{\theta}_t). $$
Consequently, in the limit of vanishing step sizes, the algorithm turns into the Itô diffusion
$$d\boldsymbol{\theta}_t = \frac{1}{2}G(\boldsymbol{\theta}_t)\nabla_{\boldsymbol{\theta}_t}u(\boldsymbol{\theta}_t)dt +\frac{1-\alpha}{2}\boldsymbol{\Gamma}(\boldsymbol{\theta}_t)dt + G(\boldsymbol{\theta}_t)^{\frac{1}{2}}d\boldsymbol{B}_t.$$
As is obvious, $\boldsymbol{\Gamma}(\boldsymbol{\theta}_t)$ is downscaled by a factor of $(1-\alpha)$ from what Riemannian Langevin dynamics would prescribe it to be.\\
The effect this has on the stationary density the process samples can be evaluated in closed form in one dimension (see Appendix A.1 for the derivation):
$$\pi(\theta)=Zp(\theta|\mathcal{D}) G(\theta)^{-\alpha}.$$
The case where $\Gamma$ was dropped entirely corresponds to the limit $\alpha \rightarrow 1.$
While in some papers the convergence was explicitly proven to be close to the stationary distribution, unfortunately, it was taken for granted that the stationary distribution is equal to the posterior while clearly the stationary distribution can deviate substantially from the posterior.
In particular, since a typical metric will become very large for vanishing gradients, this will lead to the stationary density becoming very small around local \textbf{maxima} of the target distribution where the gradients necessarily need to vanish.
\subsection{Convergence of Adam SGLD}
The Adam SGLD algorithm does not fit into the framework of Riemannian Langevin dynamics and consequently, its convergence has to be analyzed separately. 
The algorithm was given by the updates:
$$\boldsymbol{V}_{t} = \alpha \boldsymbol{V}_{t-\epsilon}+(1-\alpha) \nabla_{\boldsymbol{\theta}_t}u(\boldsymbol{\theta}_t)\odot \nabla_{\boldsymbol{\theta}_t}u(\boldsymbol{\theta}_t)$$
$$G^{\alpha,t}=diag\left(\frac{1}{\lambda + \sqrt{\boldsymbol{V}_t}}\right)$$
$$\boldsymbol{m}_t=\beta \boldsymbol{m}_{t-\epsilon} +(1-\beta)\nabla_{\boldsymbol{\theta}_t}u(\boldsymbol{\theta}_t),$$
$$\boldsymbol{\theta}_{t+\epsilon}=\boldsymbol{\theta}_{t}+\frac{\epsilon}{2}(\nabla_{\boldsymbol{\theta}_t}u(\boldsymbol{\theta}_t) +aG^{\alpha,t}\boldsymbol{m}_t)+\boldsymbol{\mathcal{N}}_t(0,\epsilon \mathbb{I}).$$
Curiously, the authors claim convergence to the posterior without any restrictions on the hyperparameter $a$ as long as the step size $\epsilon$  is sufficiently small.\\
This however can not be the case, since for sufficiently large $a$ and sufficiently small $\epsilon$ this algorithm just turns into regular Adam with step size $\frac{a \cdot \epsilon}{2}$.\\
In the limit $\epsilon \rightarrow 0$ we again have 
$$G^{\alpha,t} \rightarrow diag \left(\frac{1}{\lambda +\sqrt{\nabla_{\boldsymbol{\theta}_t}u(\boldsymbol{\theta}_t)\odot \nabla_{\boldsymbol{\theta}_t}u(\boldsymbol{\theta}_t})   }\right)=:G(\boldsymbol{\theta}_{t})$$
as well as
$$\boldsymbol{m}_t \rightarrow \nabla_{\boldsymbol{\theta}_t}u(\boldsymbol{\theta}_t).$$
Consequently, the process turns into the Itô diffusion
$$d\boldsymbol{\theta}_{t}=\frac{1}{2}\left(1+aG(\boldsymbol{\theta}_t)\right) \nabla_{\boldsymbol{\theta}_t}u(\boldsymbol{\theta}_t)dt +d\boldsymbol{B}_t.$$
Evaluating the stationary density in the one-dimensional case yields (see Appendix A.2 for the derivation):
$$\pi(\theta)=Zp(\theta|\mathcal{D}) \exp \left(\int^{\theta}aG(x)\frac{d}{dx}\log p(x|\mathcal{D})dx \right).$$
In this case, the additional factor has the opposite effect and artificially sharpens the density as will be illustrated in the example of the next section.
\section{Empirical Evaluation}
To demonstrate the correctness of the results derived in the previous section, we estimate the stationary densities of the discussed algorithms empirically in this section.\\
For PSGLD we use the official TensorFlow implementation linked on the first author's GitHub page. For SGRLD in Monge and Shampoo metric, we use the code made publically available by the authors. Because the authors don't provide an implementation where the $\boldsymbol{\Gamma}$ term is included, we run the experiments of those algorithms with the $\boldsymbol{\Gamma}$ term completely dropped.  \\
Because we could not find the source code of Adam SGLD provided by the authors, we use our own reimplementation of that algorithm. The code for running all the experiments can be found at \textbf{https://github.com/TimRensmeyer/Convergence-Experiments.git}.

\subsection{The Experimental Setup}

In this experiment, we empirically verify the derived results of the previous section, by attempting to sample from a standard normal distribution $p(\theta|\mathcal{D})=\frac{1}{\sqrt{2 \pi}} \exp \left(-\frac{\theta^2}{2} \right)$. In all experiments, we set $\alpha = 0.9$ and all stability constants to $10^{-8}$. \\
Because there is a slight deviation in the PSGLD algorithm, where in the paper the gradients of the log prior are not included in the preconditioning but in the implementation they are, we simply assume a flat prior $p(\theta)=1$ in this case. \\
The derivations for the stationary densities mentioned in the following are straightforward but somewhat tedious with the results from the last section and can be found in Appendix A.3.
Utilizing the results from the previous section and determining the normalization constants via numerical integration, we get as the stationary density for PSGLD:
$$\pi(\theta)=\frac{1.258}{\sqrt{2\pi}}exp(-\frac{1}{2}\theta^2)(10^{-8}+|\theta|)^{0.9}$$
and for SGRLD in Shampoo metric
$$\pi(\theta)=\frac{1.253}{\sqrt{2\pi}}exp(-\frac{1}{2}\theta^2)|\theta|.$$
For SGRLD in Monge metric we set $\beta=1$ and get
$$\pi(\theta)=\frac{0.5}{\sqrt{2\pi}}exp(-\frac{1}{2}\theta^2) \left( 1-\frac{x^2}{1+x^2}\right)^{-1}.$$

Finally setting $a=1$ and $\beta=0.5$ in the Adam SGLD algorithm yields the stationary density
$$\pi(\theta)=\frac{1.912}{\sqrt{2\pi}}exp(-\frac{1}{2}\theta^2)exp(-|\theta|) (|\theta|+10^{-8})^{10^{-8}}.$$\\

For $x \in [a,b)$ for a sufficiently small interval $[a,b)$ the stationary density for each algorithm can be estimated with the ergodic property via
$$\pi(x)\approx \frac{\mathbb{P}_{\pi}([a,b))}{b-a}=\frac{1}{b-a}\lim_{T\rightarrow \infty}\frac{1}{T}\int_{t=0}^T \mathbbm{1}_{[a,b)}(\theta_t)dt$$
$$\approx \frac{1}{b-a}\frac{1}{T_{max}}\int_{t=0}^{T_{max}} \mathbbm{1}_{[a,b)}(\theta_t)dt,$$
where $T_{max}$ is chosen sufficiently large. We use this estimator with intervals of width $0.1$ to estimate the stationary density by running each algorithm for $10^7$ steps at a step size of $10^{-4}$.
\subsection{Results}
\begin{figure}[h!]
\centering
\includegraphics[scale=0.63]{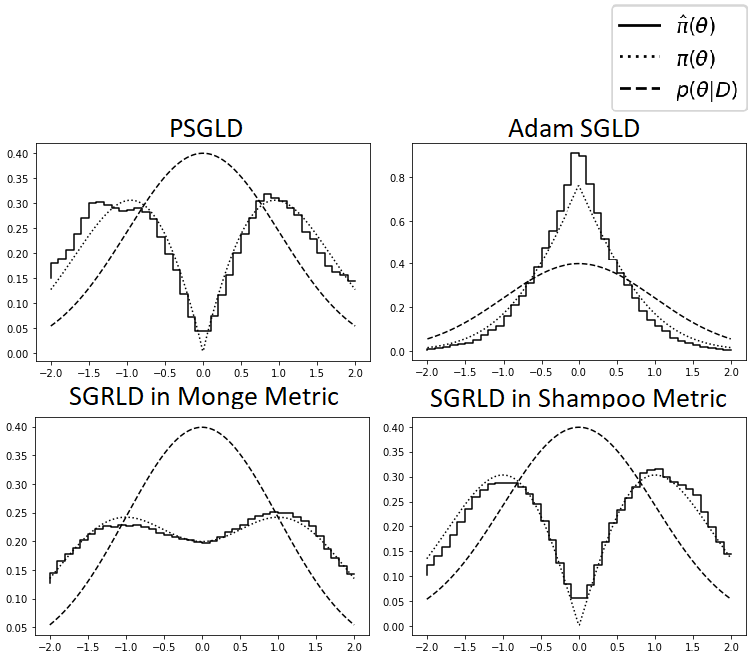}
\caption{The stationary densities when attempting to sample from a standard normal distributed posterior with each of the algorithms. $p(\theta|\mathcal{D})$ denotes the target distribution, $\pi(\theta)$ are the stationary densities we derived in the previous sections and $\hat{\pi}(\theta)$ are the empirical estimates for the stationary densities.}\label{fig}
\end{figure}
Even though the empirical estimator for the stationary density is somewhat primitive, the results in Figure \ref{fig} align well with the predictions made for the stationary densities of the algorithms derived in the previous sections and very clearly do not coincide with the target density the algorithm is supposed to converge to.

\section{Discussion and Outlook}
In this paper, we demonstrated that current approaches for making diffusion-based sampling of Bayesian neural network posteriors more scalable can introduce a substantial bias in the sampled distribution to the point where the sampled distribution has a deep local minimum at the global maximum of the target distribution. Due to their substantial bias, they clearly can not be considered sampling algorithms for the true Bayesian posterior distribution. Hence the main research question of this paper of whether the proposed algorithms converge to a distribution close to the true posterior can be conclusively answered with no they do not.
While they may still work well empirically as uncertainty quantification algorithms on some tasks, one should be aware of the biases these algorithms introduce to the sampled distribution.\\
Furthermore, PSGLD and SGRLD in Shampoo and Monge metric can actually be fixed by rescaling the $\boldsymbol{\Gamma}^{\alpha,t}$ term by 
$\frac{1}{1-\alpha}$. The resulting algorithm then has the correct stationary distribution in the limit of vanishing step sizes. However, one of the apparent main benefits, the fact that this term can be dropped completely, can not be upheld. Consequently, the updates of the fixed algorithms remain computationally demanding to compute.\\

In regards to the more open research question of what the consequences of using adaptive step sizes without a correction term are, we demonstrated that at least in one dimension, this leads to a change in the stationary distribution proportional to the inverse of the metric used. Further, we illustrated why, due to the ergodic properties of stochastic differential equations, it seems unlikely that adaptive step sizes similar to modern neural network optimizers can ever be incorporated into diffusion-based sampling methods without adding a computationally costly corrective drift term.\\

There are two related approaches to improving the convergence of diffusion-based sampling methods in the existing literature that we have not discussed so far in this paper.\\
The first one is a new approach by \citet{BatchNorm} which uses the framework of Riemannian Langevin Dynamics to imitate the batch normalization operation. Unfortunately, they also follow the erroneous argument of Li et. al for dropping the corrective drift term. Hence, the algorithm can not be expected to converge to the correct distribution although the same fix of rescaling the corrective drift term as above can be applied. Furthermore, the resulting metric depends only on the current set of parameters and not their gradient. Due to this fact, it might be possible to calculate the $\boldsymbol{\Gamma}$ term more efficiently in this case, since no second-order derivatives w.r.t. the parameters would need to be calculated.\\
However, while this approach appears promising, it seems unlikely that an imitation of batch normalization alone will result in an improved convergence speed comparable to adaptive step sizes.\\

The second approach not discussed in this paper so far is the one introduced by \citet{ColdPosterior}, who use adaptive step sizes which are not updated each time step but once every few thousand time steps. However, while this clearly would inhibit the deviation from the target distribution at finite step sizes, the limiting process one gets at vanishing step sizes is the same as the one that would result from updating the adaptive step sizes each time step. A natural question then arises, how much time there has to be between the updates of the adaptive step sizes in order for the sampled distribution to remain close to the target distribution? It appears unlikely that this time can be chosen small enough, that this procedure can be considered adapted to the local geometry around the current parameter vector. Hence it can probably not be used as a locally adaptive algorithm in any sense of the word. Wenzel et al. completely circumvent this issue by additionally using a single adaptive step size for each layer, which remains relatively constant along the entire trajectory of the process and hence is not very locally adaptive no matter how much time there is between updates of the adaptive step sizes.\\

Overall, it appears that the current approaches for making locally adaptive diffusion-based sampling methods less computationally demanding simply do not work and the ergodic theory of stochastic differential equations, as well as the results by \citet{CompleteRecipe} make it seem very difficult to make any progress in that direction. If it is achievable at all, some new ideas are clearly needed.

\bibliography{bib}
    \bibliographystyle{plainnat}
\newpage

\onecolumn

\title{Appendix}
\maketitle

\appendix

\section{Derivations}
\subsection{Derivation of the Stationary Density of Riemannian Langevin Dynamics with Downscaled $\boldsymbol{\Gamma}$ Term in One Dimension}
Utilizing the general formula for the stationary density $$\pi(\theta)=\frac{Z}{\sigma(\theta)^2}\exp\left(2\int^{\theta}\frac{\mu(x)}{\sigma(x)^2}dx\right),$$ of a one dimensional SDE $d\theta_t=\mu(\theta_t)dt +\sigma(\theta_t)dB_t$ and applying it to Riemannian Langevin Dynamics with a downscaled  $\Gamma$ term
$$d\theta_t = \frac{1}{2}G(\theta_t)\frac{d}{d\theta_t}u(\theta_t)dt +\frac{1-\alpha}{2}\Gamma(\theta_t)dt + G(\theta_t)^{\frac{1}{2}}dB_t$$
yields
$$\pi(\theta)=\frac{Z}{G(\theta)}\exp\left(2\int^{\theta} \frac{G(x)\frac{d}{dx}u(x)+(1-\alpha)\Gamma(x)}{2G(x)}dx\right),$$
$$=\frac{Z}{G(\theta)}\exp\left(\int^{\theta} \frac{G(x)\frac{d}{dx}\log p(x|\mathcal{D})+(1-\alpha)\frac{d}{dx}G(x)}{G(x)}dx\right)$$
$$=\frac{Z}{G(\theta)}\exp\left(\int^{\theta} \frac{d}{dx}\log p(x|\mathcal{D}) + \frac{d}{dx}\log(G(x)^{1-\alpha} )dx\right)$$
$$=\frac{Z}{G(\theta)}\exp\left(\log p(\theta |\mathcal{D}) + \log(G(\theta)^{1-\alpha} )\right)$$
$$=\frac{Zp(\theta |\mathcal{D}) G(\theta)^{1-\alpha} }{G(\theta)}=Zp(\theta |\mathcal{D}) G(\theta)^{-\alpha}.$$
The case in which the $\Gamma$ term was dropped completely corresponds to the limit $\alpha \rightarrow 1$.

\subsection{Derivation of the Stationary Density of Adam SGLD in One Dimension}
As was discussed, in the limit of vanishing step sizes the Adam SGLD algorithm turns into the SDE
$$d\boldsymbol{\theta}_{t}=\frac{1}{2}\left(1+aG(\boldsymbol{\theta}_t)\right) \nabla_{\boldsymbol{\theta}_t}u(\boldsymbol{\theta}_t)dt +d\boldsymbol{B}_t.$$
Using again the formula for for the stationary density for one-dimensional SDEs it follows that
$$\pi(\theta)=Z\exp \left(2\int^{\theta}\frac{1}{2}\left(1+aG(x)\right)\frac{d}{dx}u(x)dx \right)$$
$$=Z\exp \left(\int^{\theta}\left(1+aG(x)\right)\frac{d}{dx}\log p(x|\mathcal{D})dx \right)$$
$$=Zp(\theta|\mathcal{D}) \exp \left(\int^{\theta}aG(x)\frac{d}{dx}\log p(x|\mathcal{D})dx \right).$$

\subsection{Derivations of the Stationary Densities for the Experiments}

Based on the results from section 3, we expect the stationary density of all algorithms with the exception of Adam SGLD to be of the form
$$\pi(\theta)= \frac{Z}{\sqrt{2 \pi}} \exp \left(-\frac{\theta^2}{2} \right) G(\theta)^{-\alpha}.$$
where $G(\theta)$ is of the metric one gets in the limit $\epsilon \rightarrow 0$.
For PSGLD and SGRLD in Shampoo metric $G(\theta)=\frac{1}{\lambda +\sqrt{ u(\theta)^2}}=\frac{1}{\lambda +|\theta|}$ where $\lambda = 0 $ for the Shampoo metric. 
For SGRLD in Monge metric we get $G(\theta)=1-\frac{\beta ^2 u(\theta)^2}{1+\beta ^2 u(\theta)^2}=1-\frac{\beta ^2 \theta^2}{1+\beta ^2 \theta^2}$.\\

Finally, for the Adam SGLD algorithm the stationary density is 
$$\pi(\theta)=Zp(\theta|D) \exp \left(\int^{\theta}aG(x)\frac{d}{dx}\ln p(x|D)dx\right)$$
$$=Zp(\theta|D) \exp \left(a\int^{\theta}\frac{-x}{\lambda+|x|}dx \right)$$

$$=\frac{Z}{\sqrt{2 \pi}} \exp \left(-\frac{\theta^2}{2} \right) \exp \left(-a |\theta|\right)|(|\theta| +   \lambda)^{a \lambda}$$

The normalization constants $Z$ were all determined via numerical integration of the densities.

\end{document}